# Fast Dempster-Shafer Clustering Using a Neural Network Structure

**Johan Schubert**

Department of Data and Information Fusion
Division of Command and Control Warfare Technology
Defence Research Establishment
SE–172 90 Stockholm, Sweden
schubert@sto.foa.se

**Abstract:** In this article we study a problem within Dempster-Shafer theory where $2^n - 1$ pieces of evidence are clustered by a neural structure into $n$ clusters. The clustering is done by minimizing a metaconflict function. Previously we developed a method based on iterative optimization. However, for large scale problems we need a method with lower computational complexity. The neural structure was found to be effective and much faster than iterative optimization for larger problems. While the growth in metaconflict was faster for the neural structure compared with iterative optimization in medium sized problems, the metaconflict per cluster and evidence was moderate. The neural structure was able to find a global minimum over ten runs for problem sizes up to six clusters.

## 1   Introduction

In this article we will study a neural structure for clustering evidence in large scale problems within Dempster-Shafer theory [12]. The studied problem concerns the situation when we are reasoning with multiple events which should be handled independently. We use the clustering process to separate the evidence into subsets that will be handled separately.

In earlier work [4–9] we developed a method based on iterative optimization for the clustering of evidence in medium sized problems. That method was developed as a part of a multiple-target tracking algorithm for an antisubmarine intelligence analysis system [1–2]. In a subsequent paper [10] we developed a classification method for incoming pieces of evidence. Here, we used prototypes in order to obtain faster classification. These prototypes were derived from a previous clustering process. That method further increased the computation speed of the iterative optimization for small and medium sized problems, but it did little for larger problems.

For large scale problems it became clear that we need a method with much lower computational complexity. To achieve this we are prepared to sacrifice some of the clustering performance, if necessary.

The solution described in this article is based on clustering with a neural structure. We will use a neural network, but we will not do any learning to set the weights of the network. Instead, all the weights will be directly set by a method where we use the conflict in Dempster's rule as input to setting the weights.

420

In a recent paper [11] this method was further extended for simultaneous clustering and determination of number of clusters during iteration in the neural structure. Here, we let the output signals of neurons represent the degree to which a pieces of evidence belong to a corresponding cluster. From these we derive a probability distribution regarding the number of clusters, which gradually during the iteration is transformed into a determination of number of clusters. This gradual determination is fed back into the neural structure at each iteration to influence the clustering process.

Many of the ideas in this article are inspired by a solution to the traveling salesman problem by Hopfield and Tank [3]. They used a neural network as an effective method to find a good shortest path between several cities.

In Section 2 we describe the problem at hand and in Section 3 we give an overview of the iterative optimization solution developed in [4]. Then we describe the neural structure to achieve effective clustering (Section 4). We end by presenting a comparison between the neural structure and iterative optimization (Section 5).

## 2 The problem

If we receive several pieces of evidence about different and separate events and the pieces of evidence are mixed up, we want to arrange the them according to which event they are referring to. Thus, we partition the set of all pieces of evidence $\chi$ into subsets where each subset refers to a particular event. In figure 1 these subsets are denoted by $\chi_i$ and the conflict when all pieces of evidence in $\chi_i$ are combined by Dempster's rule is denoted by $c_i$. Here, thirteen pieces of evidence are partitioned into four subsets. When the number of subsets is uncertain there will also be a "domain conflict" $c_0$ which is a conflict between the current hypothesis about the number of subsets and our prior belief. The partition is then simply an allocation of all pieces of evidence to the different events. Since these events do not have anything to do with each other, we will analyze them separately.

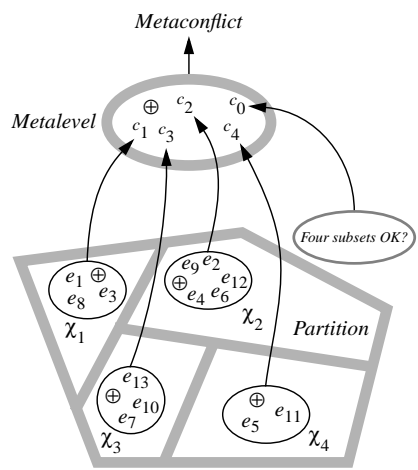

Figure 1: The conflict in each subset of the partition becomes a piece of evidence at the metalevel

Now, if it is uncertain to which event some pieces of evidence is referring we have a problem. It could then be impossible to know directly if two different pieces of evidence are referring to the same event. We do not know if we should put them into the same subset or not. This problem is then a problem of organization. Evidence from different events that we want to analyze are unfortunately mixed up and we are facing a problem in separating them.

To solve this problem, we can use the conflict in Dempster's rule when all pieces of evidence within a subset are combined, as an indication of whether these pieces of evidence belong together. The higher this conflict is, the less credible that they belong together.

Let us create an additional piece of evidence for each subset with the proposition that this is not an "adequate partition". We have a simple frame of discernment on the metalevel $\Theta = \{ AdP, \neg AdP \}$, where AdP is short for "adequate partition." Let the proposition take a value equal to the conflict of the combination within the subset,

$$m_{\chi_i}(\neg AdP) \triangleq \text{Conf}(\{e_j | e_j \in \chi_i\}).$$

These new pieces of evidence, one regarding each subset, reason about the partition of the original evidence. Just so we do not confuse them with the original evidence, let us call this evidence "metalevel evidence" and let us say that its combination and the analysis of that combination take place on the "metalevel," figure 1.

We establish [4] a criterion function of overall conflict called the metaconflict function for reasoning with multiple events. The metaconflict is derived as the plausibility that the partitioning is correct when the conflict in each subset is viewed as a piece of metalevel evidence against the partitioning of the set of evidence, $\chi$, into the subsets, $\chi_i$.

DEFINITION. *Let the* metaconflict function,

$$Mcf(r, e_1, e_2, ..., e_n) \triangleq 1 - (1 - c_0) \cdot \prod_{i=1}^{r} (1 - c_i),$$

*be the conflict against a partitioning of n evidences of the set $\chi$ into r disjoint subsets $\chi_i$. Here, $c_i$ is the conflict in subset i and $c_0$ is the conflict between r subsets and propositions about possible different number of subsets.*

We will use the minimizing of the metaconflict function as the method of partitioning the evidence into subsets representing the events. This method will also handle the situation when the number of events are uncertain.

The method of finding the best partitioning is based on an iterative minimization of the metaconflict function. In each step the consequence of transferring a piece of evidence from one subset to another is investigated.

After this, each subset refers to a different event and the reasoning can take place with each event treated separately.

## 3 Iterative optimization

For a fixed number of subsets a minimum of the metaconflict function can be found by an iterative optimization among partitionings of evidences into different subsets.

In each step of the optimization the consequence of transferring evidence from one subset to another is investigated. If a piece of evidence $e_q$ is transferred from $\chi_i$ to $\chi_j$ then the conflict in $\chi_j$, $c_j$, increases to $c_j^*$ and the conflict in $\chi_i$, $c_i$, decreases to $c_i^*$. Given this, the metaconflict is changed to

$$Mcf^* = 1 - (1-c_0) \cdot (1-c_i^*) \cdot (1-c_j^*) \cdot \prod_{k \neq i,j}(1-c_k)$$
$$= 1 - (1-c_0) \cdot \prod_k (1-c_k).$$

The transfer of $e_q$ from $\chi_i$ to $\chi_j$ is favorable if Mcf* < Mcf. This is the case if

$$\frac{1-c_j^*}{1-c_j} > \frac{1-c_i}{1-c_i^*}.$$

It is, of course, most favorable to transfer $e_q$ to $\chi_k$, $k \neq i$, where Mcf* is minimal.

When several different pieces of evidence may be favorably transferred it will be most favorable to transfer the evidence $e_q$ that minimizes Mcf*.

It should be remembered that this analysis concerns the situation where only one piece of evidence is transferred from one subset to another. It may not be favorable at all to simultaneously transfer two or more pieces of evidence which are deemed favorable for individual transfer.

The algorithm, like all hill-climbing–like algorithms, guarantees finding a local but not a global optimum.

## 4 Neural structure

We will study a series of problems where $2^n - 1$ pieces of evidence, all simple support functions with elements from $2^\Theta$, are clustered into $n$ clusters, where $\Theta = \{1, 2, 3, ..., n\}$.

Thus, there is always a global minimum to the metaconflict function equal to zero, since we can take all pieces of evidence that includes the 1–element and put them into cluster 1, of the remaining evidence take all those that includes the 2–element and put them into subset 2, and so forth. Since all evidence of cluster 1 includes the 1–element there intersection is nonempty, and all evidence of cluster 2 includes the 2–element their intersection is also nonempty, etc. Thus, all conflicts $c_i$ are zero and we will always have a global minimum with Mcf = 0. This makes it easy to use Mcf as a standard for the efficiency of the clustering process.

The reason we choose a problem where the minimum metaconflict is zero is that it makes a good test example for evaluating performance. If another problem had been used we would have no knowledge of the global minimum and evaluation would be more difficult. We have no reason to believe that this choice of test examples is atyp-ical with respect to network performance.

We will choose an architecture that minimizes a sum. Thus, we have to make some change to the function that we want to minimize. If we take the logarithm of one minus the metaconflict function, we can change from minimizing Mcf to minimizing a sum.

Let us change the minimization as follows

$$\min Mcf = \min 1 - \prod_i (1 - c_i)$$
$$\max 1 - Mcf = \max \prod_i (1 - c_i)$$
$$\max \log(1 - Mcf) = \max \log \prod_i (1 - c_i)$$
$$= \max \sum_i \log(1 - c_i) = \min \sum_i -\log(1 - c_i)$$

where $-\log(1 - c_i) \in [0,\infty]$ is a weight [12, p. 77] of evidence, i.e., metaconflict.

Since the minimum of Mcf (= 0) is obtained when the final sum is minimal (= 0) the minimization of the final sum yields the same result as a minimization of Mcf would have.

Thus, in the neural network we will not let the weights be directly dependent on the conflicts between different pieces of evidence but rather on $-\log(1 - c_{jk})$, where $c_{jk}$ is the conflict between the $j$th and $k$th piece of evidence;

$$c_{jk} = \begin{cases} m_j \cdot m_k, & \text{conflict} \\ 0, & \text{no conflict.} \end{cases}$$

This, however, is a slight simplification since the neural structure will now minimize a sum of $-\log(1 - c_{jk})$, but take no account of higher order terms in the conflict. The actual function being minimized is

$$\sum_i \sum_{\substack{k,l \\ e_k, e_l \in \chi_i}} -\log(1 - c_{kl}) = \sum_i -\log \prod_{\substack{k,l \\ e_k, e_l \in \chi_i}} (1 - c_{kl})$$
$$= \sum_i -\log\left(1 - \left[\sum_{\substack{k,l \\ e_k, e_l \in \chi_i}} c_{kl} - Y\right]\right)$$

while the function above can be rewritten as

$$\sum_i -\log(1 - c_i) = \sum_i -\log\left(1 - \left[\sum_{\substack{k,l \\ e_k, e_l \in \chi_i}} c_{kl} - X\right]\right)$$

where X and Y are the higher order terms.

These functions are identical in there first order terms, and Y ≤ X. Thus, the actual minimization slightly overestimates the conflict within the subset. This is the price we have to pay to achieve fast clustering.

Let us now study the calculations taking place in the neural network during an iteration. We will use the same terminology as Hopfield and Tank [3] with input voltages as the weighted sum of input signals to a neuron, output voltages as the output signal of a neuron, and inhibition terms as negative weights.

For each neuron $n_{mn}$ we will calculate an input voltage u as the weighted sum of all signals from row $m$ and column $n$, figure 2.

This sum is the previous input voltage of the previous iteration for $n_{mn}$ plus a gain factor times the sum of the weighted sum of output voltages $V_{ij}$ of all neurons of the same column or row as $n_{mn}$ plus an excitation bias and minus the previous input voltage of $n_{mn}$.

In the column the output voltages are weighted by a data-term inhibition times the weight of conflict plus a global inhibition;

$$\sum_i (-\text{dti} \cdot \log(1 - c_{in}) + \text{gi}) \cdot V_{in}$$

where dti is the data-term inhibition, gi the global inhibition, $V_{in}$ is the output voltage from neuron $n_{in}$, and $i$ is an index over all rows of the column.

In the row the $V_{mj}$'s are weighted by the sum of row inhibition and global inhibition;

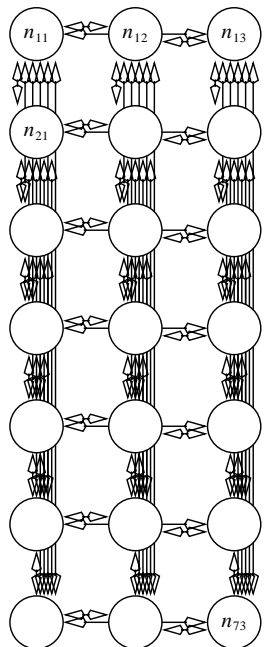

Figure 2: Neural network. Each column corresponds to a cluster and each row corresponds to a piece of evidence

$$\sum_{j \neq n} (\text{ri} + \text{gi}) \cdot V_{mj}$$

where ri is the row inhibition, and $j$ an index over all columns of the row.

Thus, the new input voltage to $n_{mn}$ at iteration $t+1$ is

$$u_{mn}^{t+1} = u_{mn}^{t} + \eta \cdot \left( \sum_{i} [-\text{dti} \cdot \log(1 - c_{in}) + \text{gi}] \cdot V_{in} \right.$$
$$\left. + \sum_{j \neq n} [\text{ri} + \text{gi}] \cdot V_{mj} + \text{eb} - u_{mn}^{t} \right)$$

where $\eta$ is the gain factor.

In the experiments we used the following parameter settings: $\eta = 10^{-5}$, ri = −500. Initially dti was set at −2000 and gi was set at −200. Both were lowered as the problem size grew. This is to assure that the inhibitory signals from the column does not overwhelm the signals from the row as the column length grows like $2^n$ while the row length grows like $n$ as the problem size $n$ grows. The excitation bias was set at

$$\text{eb} = \frac{-[\text{gi} \cdot (2^n - 1) + (\text{ri} + \text{gi}) \cdot (n - 1)]}{n}$$

where $n$ is the number of columns, i.e., clusters.

The task of parameter fine tuning grows with the size of the neural network. For larger problem it might be necessary to do this thing automatically, although this has not been done here.

Finally, from the new input voltage to $n_{mn}$ we can calculate a new output voltage of $n_{mn}$

$$V_{mn}^{t+1} = \frac{1}{2} \cdot \left( 1 + \tanh\left(\frac{u_{mn}^{t+1}}{u_0}\right) \right)$$

where tanh is the hyperbolic tangent, $u_0 = 0.02$, and $V_{mn}^{t+1} \in [0,1]$.

Initially, before the iteration begins, each neuron is initiated with an input voltage of $u_{00}$ + noise where

$$u_{00} = u_0 \cdot \text{atanh}\left(\frac{2}{n} - 1\right)$$

and atanh is the hyperbolic arc tangent.

The initial input voltage is set at $u_{00} + \delta u$ where $\delta u$, the noise, is a random number chosen uniformly in the interval $-0.1 \cdot u_0 \leq \delta u \leq 0.1 \cdot u_0$.

In each iteration all new voltages are calculated from the results of the previous iteration. This continues until convergence is reached. As long as the weights of the neural network is symmetric convergence is always guaranteed. This is always the case here since the only factor that varies is the conflict between two pieces of evidence. Thus, the weights from $n_{in}$ to $n_{jn}$ and from $n_{jn}$ to $n_{in}$ are equal.

In each iteration we need to make some special checks.

First, we assure that not all output voltages of a row of neurons decrease during the same iteration. That could possibly lead to the piece of evidence corresponding to that row not being clustered at all. If this happens we will add to all output voltages of the row so that the one that decreased the least is now unchanged.

This control plus the fact that we have logical conditions only on the row, and data-terms from only one column for each neuron makes our problem easier than Hopfield and Tank's model for the traveling salesman problem. They had logical conditions on both row and column plus data-terms from both the previous and next columns for each neuron. This allows us to avoid the problems with convergence and performance, as described by Wilson and Pawley [13], of the Hopfield and Tank model.

Secondly, we check the two highest output voltages of the row. If the highest output voltage is greater or equal to 0.99 then it is set to 1.0 and all other output voltages of the row are set to 0.0, or if the second highest output voltage is 0.0, then regardless of the value of the highest output voltage, the highest output voltage is set to 1.0. This is done merely to speed up convergence.

In figure 3 the convergence of a 155-neural network with 31 rows and five columns for clustering 31 pieces of evidence into five subset is shown. This leads here to a

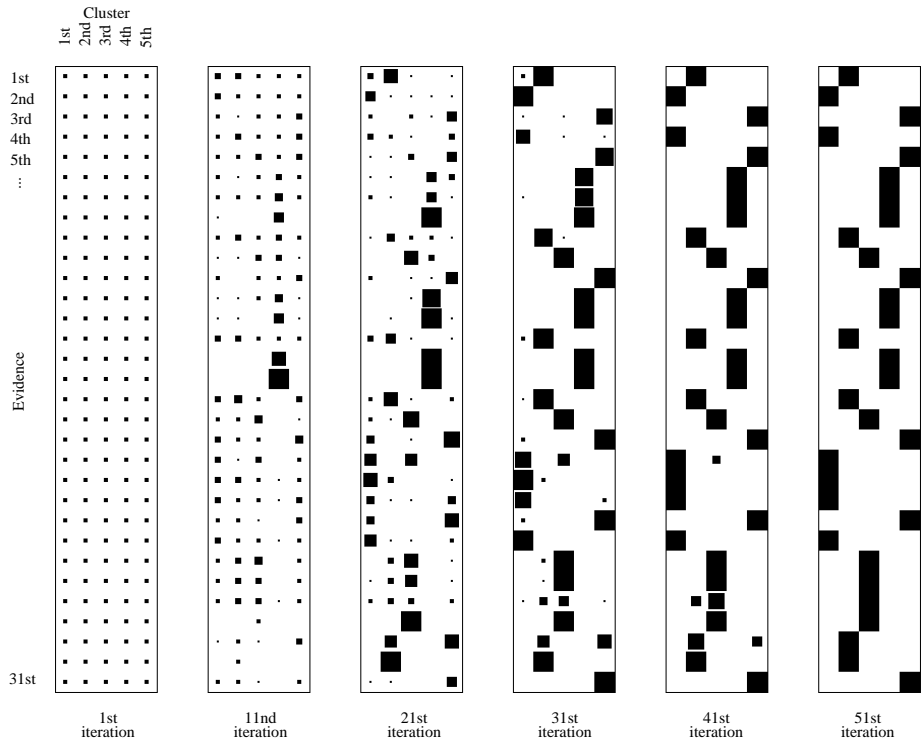

Figure 3: Six different states (iterations) of a neural network with 155 neurons. From left to right: The convergence of clustering 31 pieces of evidence into five clusters at the first, eleventh, 21st, 31st, 41st, and 51st (final state) iteration. In each snap-shot of an iteration each of the five columns represent one cluster and each of the 31 rows represent one piece of evidence. The linear dimension of each square is proportional to the output voltage of the neuron and represent the degree to which a pieces of evidence belong to a cluster. In the final state each row has one output voltage of 1.0 and four output voltages of 0.0. A piece of evidence, represented by a row, is now clustered into the cluster where the output voltage is 1.0

global optimum being found in 51 iterations.

After convergence is achieved the conflict within each cluster, i.e., column, is calculated by combining those pieces of evidence for which the output voltage for the column is 1.0.

We now have a conflict for each subset and can calculate the overall metaconflict, Mcf, by the previous formula.

## 5 Results

In this section we investigate the clustering performance and computation time of the two clustering processes for the neural structure and the iterative optimization. We make this comparison as the problem size grows.

In all problem sizes we will try clustering $2^n - 1$ pieces of evidence into $n$ subsets. As reported before the evidence support all different subsets of the frame $\Theta = \{1, 2, 3, ..., n\}$. Thus, we know that the metaconflict function has a global minimum with metaconflict equal to zero.

In figure 4 we notice that the iterative optimization has an exponential computation time in the number of items of evidence. The neural structure has a much lower complexity although it has a higher computation time for small problems. For problems up to five subsets and 31 pieces of evidence the iterative optimization is the fastest, but from six subsets and 63 pieces of evidence the neural structure is vastly superior. Now let us study the clustering performance: Will we find a global optimum?

Yes, we find that the best run out of ten different runs in both methods manages to find a global optimum for all problem sizes of three to six subsets. However, we also notice that the median and mean minimum metaconflict are much higher for the neural structure than for the iterative optimization.

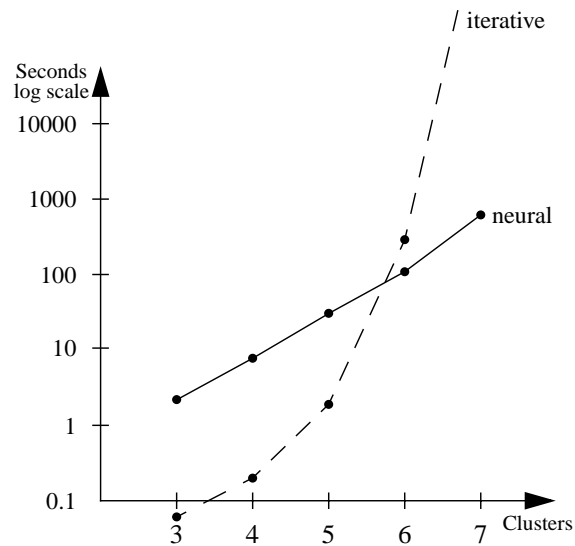

Figure 4: Computation time (mean of 10 runs) of neural structure compared to iterative optimization

For the neural structure we have mean conflicts of 0.016, 0.059, 0.076, 0.398 and 0.856 for problem sizes three to seven, while for iterative optimization the mean conflicts are 0, 0.001, 0.003 and 0.097 for problem sizes three to six.

How serious is this metaconflict?

If we make an assumption that in a local minimum the different conflicts from the clusters are equal, they certainly are at a global minimum (equal to zero), we may then calculate the average conflict of a cluster.

With $c_0 = 0$ and $c_i = c_j$ for all $i, j$ we have

$$c_i = 1 - (1 - Mcf)^{1/n}.$$

We see in figure 5 that it grows much slower than the total metaconflict. Thus, a large part of the growth in metaconflict depends on the increased number of clusters whose conflicts becomes additional terms in the metaconflict function.

Notice that in the six cluster problem we have roughly ten pieces of evidence in each cluster. These pieces of evidence have on average a basic probability number of 0.5. Still the median conflict in a cluster is only 0.094.

To investigate this performance a bit closer still, let us first study the probability of a conflict between different pieces of evidence.

With $2^n - 1$ pieces of evidence, all simple support functions with elements from the set of all subsets of $\Theta = \{1, 2, 3, ..., n\}$, there are

$$\frac{1}{2} \cdot ((2^n - 1)^2 - (2^n - 1))$$

possible combinations.
Of these

$$\frac{1}{2} \cdot \sum_{j=1}^{n-1} \binom{n}{j} \cdot \sum_{k=1}^{n-j} \binom{n-j}{k}$$

are in conflict.

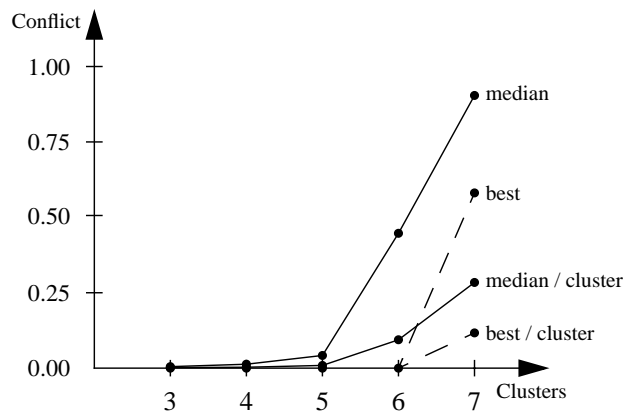

Figure 5: Conflict and conflict per cluster

If we draw two different random pieces of evidence from the set of all subsets we have a probability of conflict between there propositions of

$$P(\text{Conf}(m_i, m_j | i \neq j)) = \frac{\sum_{j=1}^{n-1} \binom{n}{j} \cdot \sum_{k=1}^{n-j} \binom{n-j}{k}}{(2^n - 1)^2 - (2^n - 1)}, \quad 2^n - 1 = |\Theta|,$$

where, e.g., $P(\text{Conf}(m_i, m_j | i \neq j)) = 0.152$ when $n = 6$.

We may compare the total median conflict in a cluster of 0.094 with the average conflict between two known conflicting pieces of evidence of 0.25, or the average conflict between two random selected pieces of evidence of 0.038 (by using the formula above).

In an local minimum the probability is much smaller since the pieces of evidence are clustered to avoid other conflicting pieces of evidence.

When the number of misplaced pieces of evidence are close to zero it might be relevant to measure the median metaconflict per cluster and evidence. (We already found that local optima were good with very few misplaced pieces of evidence).

In figure 6 below, we see that the median metaconflict per cluster and evidence is quite moderate, although it grows in the six and seven cluster problem.

Let us study the best clustering of the seven cluster problem. We find conflicts of 0.05, 0, 0.21, 0.17, 0, 0 and 0.32, respectively, in the seven different clusters. The median conflict is 0.05 in the 1st cluster. The 1st cluster contains 17 pieces of evidence. All but two of them contains the 6-element. The two remaining support {4, 5, 7} and {1, 2, 5, 7}, respectively. At least one of these elements are also present in all but one other piece of evidence. The two pieces of evidence {4, 5, 7} and {1, 2, 5, 7} are only in conflict with {6}.

Thus, of the 136 pairs of evidence in the 1st cluster only two pairs have a conflict. This is a small price to pay to obtain effective clustering. Had we chosen 17 random selected items we could have expected 16.4 of the 136 pairs to be in conflict (by the formula above).

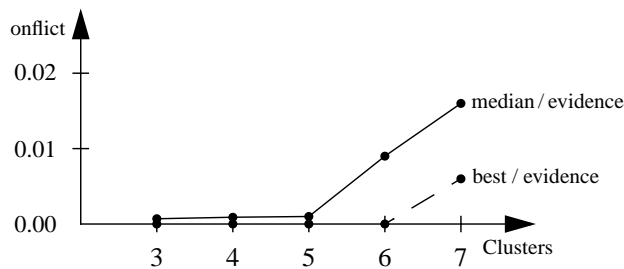

Figure 6: Median conflict per evidence

## 6  Conclusions

We have demonstrated that a neural structure is effective for clustering evidence in large scale problems. In the trials with $2^n - 1$ pieces of evidence clustered into *n* clusters the neural structure was faster than iterative optimization for problems when clustering 63 pieces of evidence into six clusters or larger. While the best of ten runs found a global optimum for both methods for all problem sizes up to six clusters the median metaconflict was higher for the neural structure. However, since a good best run was found and the median conflict per cluster and evidence was moderate, this was deemed acceptable.

## References

[1] U. Bergsten, and J. Schubert, Dempster's rule for evidence ordered in a complete directed acyclic graph, *Int. J. Approx. Reason.* 9(1), 37–73, 1993.

[2] U. Bergsten, J. Schubert, and P. Svensson, Applying data mining and machine learning techniques to submarine intelligence analysis, in *Proc. 3rd Int. Conf. Knowledge Discovery and Data Mining,* pp. 127–130, Newport Beach, August 1997.

[3] J.J. Hopfield, and D.W. Tank, "Neural" Computation of Decisions in Optimization Problems, *Biol. Cybern.* 52, 141–152, 1985.

[4] J. Schubert, On Nonspecific Evidence, *Int. J. Intell. Syst.* 8(6), 711–725, 1993.

[5] J. Schubert, Cluster-based Specification Techniques in Dempster-Shafer Theory for an Evidential Intelligence Analysis of Multiple Target Tracks, Ph.D. Thesis, TRITA–NA–9410, Royal Institute of Technology, Sweden, 1994, ISRN KTH/NA/R—94/10—SE, ISSN 0348–2952, ISBN 91–7170–801–4.

[6] J. Schubert, Cluster-based Specification Techniques in Dempster-Shafer Theory, in *Symbolic and Quantitative Approaches to Reasoning and Uncertainty,* pp. 395–404. Springer-Verlag (LNAI 946), Berlin, 1995.

[7] J. Schubert, Cluster-based Specification Techniques in Dempster-Shafer Theory for an Evidential Intelligence Analysis of Multiple Target Tracks (Thesis Abstract), *AI Comm.* 8(2), 107–110, 1995.

[8] J. Schubert, Finding a Posterior Domain Probability Distribution by Specifying Nonspecific Evidence, *Int. J. Uncertainty, Fuzziness and Knowledge-Based Syst.* 3(2), 163–185, 1995.

[9] J. Schubert, Specifying Nonspecific Evidence, *Int. J. Intell. Syst.* 11(8), 525–563, 1996.

[10] J. Schubert, Creating Prototypes for Fast Classification in Dempster-Shafer Clustering, in *Qualitative and Quantitative Practical Reasoning,* pp. 525–535. Springer-Verlag (LNAI 1244), Berlin, 1997.

[11] J. Schubert, Simultaneous Dempster-Shafer Clustering and Gradual Determination of Number of Clusters Using a Neural Network Structure, in *Proc. 1999 Information, Decision and Control,* pp. 401–406, Adelaide, February 1999.

[12] G. Shafer, *A Mathematical Theory of Evidence,* Princeton University Press, Princeton, 1976.

[13] G.V. Wilson, and G.S. Pawley, On the Stability of the Traveling Salesman Problem Algorithm of Hopfield and Tank, *Biol. Cybern.* 58, 63–70, 1988.